\documentclass[11pt,a4paper]{article}
\usepackage[hyperref]{acl2020}
\usepackage{times}
\usepackage{latexsym}

\usepackage{microtype}

\usepackage{url}
\usepackage{kotex}

\usepackage{graphicx}
\usepackage{amsmath}
\usepackage{multirow}
\usepackage{url}
\usepackage{subcaption}
\usepackage[ruled]{algorithm2e}
\usepackage{kotex}
\usepackage{tabularx}
\usepackage{caption}
\usepackage{esvect}
\usepackage{booktabs}
\usepackage{bm}
\usepackage{fdsymbol}

\usepackage{array}
\usepackage{xcolor}

\usepackage{adjustbox}
\usepackage{enumitem}

\usepackage{rotating}

\aclfinalcopy % Uncomment this line for the final submission
\def\aclpaperid{3291} %  Enter the acl Paper ID here

\title{Speaker Sensitive Response Evaluation Model}

\author{JinYeong Bak \\
  School of Computing \\
  KAIST \\
  {\tt jy.bak@kaist.ac.kr} \\\And
  Alice Oh \\
  School of Computing \\
  KAIST \\
  {\tt alice.oh@kaist.edu} \\}

\date{}

\begin{document}
\maketitle
\begin{abstract}
  Automatic evaluation of open-domain dialogue response generation is very challenging because there are many appropriate responses for a given context.  
Existing evaluation models merely compare the generated response with the ground truth response and rate many of the appropriate responses as inappropriate if they deviate from the ground truth.
One approach to resolve this problem is to consider the similarity of the generated response with the conversational context.
In this paper, we propose an automatic evaluation model based on that idea and learn the model parameters from an unlabeled conversation corpus.
Our approach considers the speakers in defining the different levels of similar context.
We use a Twitter conversation corpus that contains many speakers and conversations to test our evaluation model.
Experiments show that our model outperforms the other existing evaluation metrics in terms of high correlation with human annotation scores.
We also show that our model trained on Twitter can be applied to movie dialogues without any additional training.
We provide our code and the learned parameters so that they can be used for automatic evaluation of dialogue response generation models.

\end{abstract}

\section{Introduction}
\label{sec:introduction}

\begin{table*}[t]
    \small
    \centering
    \begin{tabular}{llrrrrrr}

        \multicolumn{2}{r}{\multirow{2}{*}{Context}}              & \multicolumn{6}{l}{\textit{A}: What do you want to do tonight?}    \\
        \multicolumn{2}{l}{} & \multicolumn{6}{l}{\textit{B}: Why don't we go see a movie?}                 \\
        \multicolumn{2}{r}{Ground truth response} & \multicolumn{6}{l}{\textit{A}: Yeah Let's go to the theater} \\

        \toprule
\multicolumn{2}{c}{Utterance}                   & \multicolumn{1}{c}{BLEU} & \multicolumn{1}{c}{ROUGE} & \multicolumn{1}{c}{EMB} & RUBER & SSREM & \multicolumn{1}{c}{Human} \\
        \midrule
A1                           & That sounds good! Have you seen Thor?          & 0.00 (3)                       & 0.00 (3)                     & 0.95 (2)                       &  0.59 (2) &   0.64 (1) & 5.00 (1)                     \\
A2                           & Good, What movie?                              & 0.00 (3)                      & 0.00 (3)                    & 0.92 (4)                      &  0.55 (4) &  0.62 (2) & 5.00 (1)                     \\
A3                           & Or hang out in city                            & 0.00  (3)                     & 0.00  (3)                    & 0.89 (6)                       &  0.48 (5) &   0.49 (3) & 3.80 (3)                     \\
N1                           & The weather is no good for walking             & 0.32 (1)                      & 0.15 (2)                     & 0.94 (3)                       &  0.47 (6) &   0.44 (4)  & 2.60 (4)                     \\
N2                           & The sight is extra beautiful here              & 0.32 (1)                      & 0.17 (1)                     & 0.97 (1)                            & 0.64 (1) & 0.38 (5) & 1.00 (5)                    \\
N3                           & Enjoy your concert              & 0.00 (3)                      & 0.00 (3)                     & 0.91 (5)                                       & 0.57 (3) & 0.33 (6) & 1.00 (5)                     \\
        \bottomrule
        \end{tabular}
    \caption{Example of appropriate responses (A1 - A3) and non-appropriate responses (N1 - N3) for a given context and ground truth response, and the responses' scores by evaluation metrics. 
    Emb is embedding average and Human is average scores from five people.
    Ranks are shown in brackets.
    SSREM has positive correlation with human scores.
    }
    \label{tab:Metric_conv_response_example}
\end{table*}

Evaluating the system generated responses for open-domain dialogue is a difficult task. There are many possible appropriate responses given a dialogue context, and automatic metrics such as BLEU \cite{P02-1040} or ROUGE \cite{W04-1013} rate the responses that deviate from the ground truth as inappropriate. Still, it is important to develop and use an automatic metric because human annotation is very costly.
In addition to BLEU and ROUGE, there is a widely-used evaluation metric based on the distributed word representation \cite{D16-1230}, but this metric shows low correlations with human judgments.

One reason for the difficulty in developing an automatic metric that correlates well with human judgements is that the range of appropriate responses for a given context is very wide. 
Table \ref{tab:Metric_conv_response_example} shows an example of a conversation between Speaker \textit{A} and \textit{B}.
While there is a ground truth response ``Yeah let's go to the theater," \textit{A} could have also said ``That sounds good! Have you seen Thor?" or ``Good. What movie?" Note that based on word overlap with the ground truth, these two responses would receive low scores. Responses labeled N\#, such as ``The weather is no good for walking" are not appropriate.
As the Table shows, the existing metrics from BLEU to RUBER are not able to tell apart these appropriate A\# responses from the inapproriate N\# responses. 

Some recent metrics such as ADEM \cite{lowe-etal-2017-towards} and RUBER \cite{tao2018ruber} compute the similarity between a context and a generated response.
However, ADEM requires human-annotated scores to train and thus cannot be applied to new datasets and domains. 
RUBER overcomes this limitation by using the idea that a random response should be used as a ``negative sample", but it is not able to distinguish the responses in the example in Table \ref{tab:Metric_conv_response_example},
because it uses only one random sample which does not provide sufficient information about appropriate and inappropriate responses.

In this paper, we propose Speaker Sensitive Responses Evaluation Model (SSREM) that analyzes the appropriateness of the responses.
We use speaker sensitive responses that are generated by one speaker to train the model.
We test SSREM in comparison with other evaluation metrics.
First, we make annotated human scores for responses in Twitter conversation data.
The evaluation scores of SSREM shows a higher correlation with human scores than other evaluation metrics.
And SSREM outperforms other metrics in terms of identifying the ground truth responses given a context.
We show the additional advantage of SSREM:
it can be applied to evaluate a new corpus in a different domain.
We train SSREM on Twitter corpus and test it on a corpus of movie reviews, and we show that SSREM outperforms other metrics in terms of the correlation with human scores and the task of identifying the ground truth response.

Our contributions in this paper include the following.
\begin{itemize}
    \item We present SSREM, a new response evaluation model trained with speaker sensitive negative samples (Sec \ref{sec:sssrem}).
    \item
    We conduct experiments on a Twitter conversation corpus and show that SSREM outperforms the others (Sec \ref{sec:exp1_results} and \ref{sec:exp2}). We further show the applicability of SSREM with Movie dialogue corpus that are not using in the training (Sec \ref{sec:exp3}).
\item We provide our code and the learned parameters of SSREM which can be used for evaluation of generated responses\footnote{\url{https://github.com/NoSyu/SSREM}}.
\end{itemize}

\section{Related Work}
\label{sec:relatedwork}
In this section, we describe existing automatic evaluation metrics for dialogue response generation and discuss their limitations.

For task-oriented dialogue models such as airline travel information system \cite{tur2010left}, completing the given task is most important, and the evaluation metrics reflect that \cite{hastie2012metrics,bordes2016learning}.
But open-domain conversation models do not have specific assigned tasks; the main goal of an open-domain conversation model is generating appropriate responses given a conversation about any topic.

Existing automatic evaluation metrics compare a generated response and the ground truth response.
The most widely-used metric are BLEU \cite{P02-1040} and ROUGE \cite{W04-1013} based on the overlap of words between the two responses.
A limitation of these word overlap-based metrics is that they cannot identify the synonyms, 
and to overcome this limitation, the embedding-based metrics use distributed word vector representations \cite{D16-1230}.
However, these metrics have poor correlation with human judgments \cite{D16-1230,novikova-etal-2017-need,gupta-etal-2019-investigating} because they still only look at the similarity between the generated response and the ground truth.
SSREM is a model with the awareness that a response can be different from the ground truth response but still appropriate for the conversation context.

The responses for a casual conversation can be varied.
For example, there are four appropriate responses including ground truth response for a given context in Table \ref{tab:Metric_conv_response_example}.
Some previous approaches suggest considering the context together with the response such as ADEM \cite{lowe-etal-2017-towards} and RUBER \cite{tao2018ruber}.
ADEM uses pre-trained VHRED \cite{serban2017hierarchical} to encode the texts and compute the score by mixing similarities among the context, generated response and a ground truth.
One limitation of ADEM is that it requires human annotated scores to learn the model.
Human labeling is cost-intensive, so it is impractical to apply to a new dataset or domain.
RUBER uses negative sampling to overcome this issue, but it uses only one random negative sample against one positive sample which is not ideal \cite{pmlr-v9-gutmann10a}.
SSREM does not require human scores to learn the model and uses many speaker sensitive negative samples.

\section{Speaker Sensitive Response Evaluation Model}
\label{sec:sssrem}
This section describes our Speaker Sensitive Response Evaluation Model (SSREM) that trains with speaker sensitive utterance samples.
SSREM looks at a given context and its ground truth response together to evaluate a generated response.
We describe the motivation of SSREM with empirical observations in section \ref{sec:SSREM_motivation}.
We present the structure of SSREM in section \ref{sec:SSREM_structure}.
With the motivation, we present a training method of SSREM with speaker sensitive utterance samples in section \ref{sec:SSREM_training}.

\subsection{Motivation}
\label{sec:SSREM_motivation}

\begin{figure}[t]
    \centering
    \includegraphics[width=0.48\textwidth]{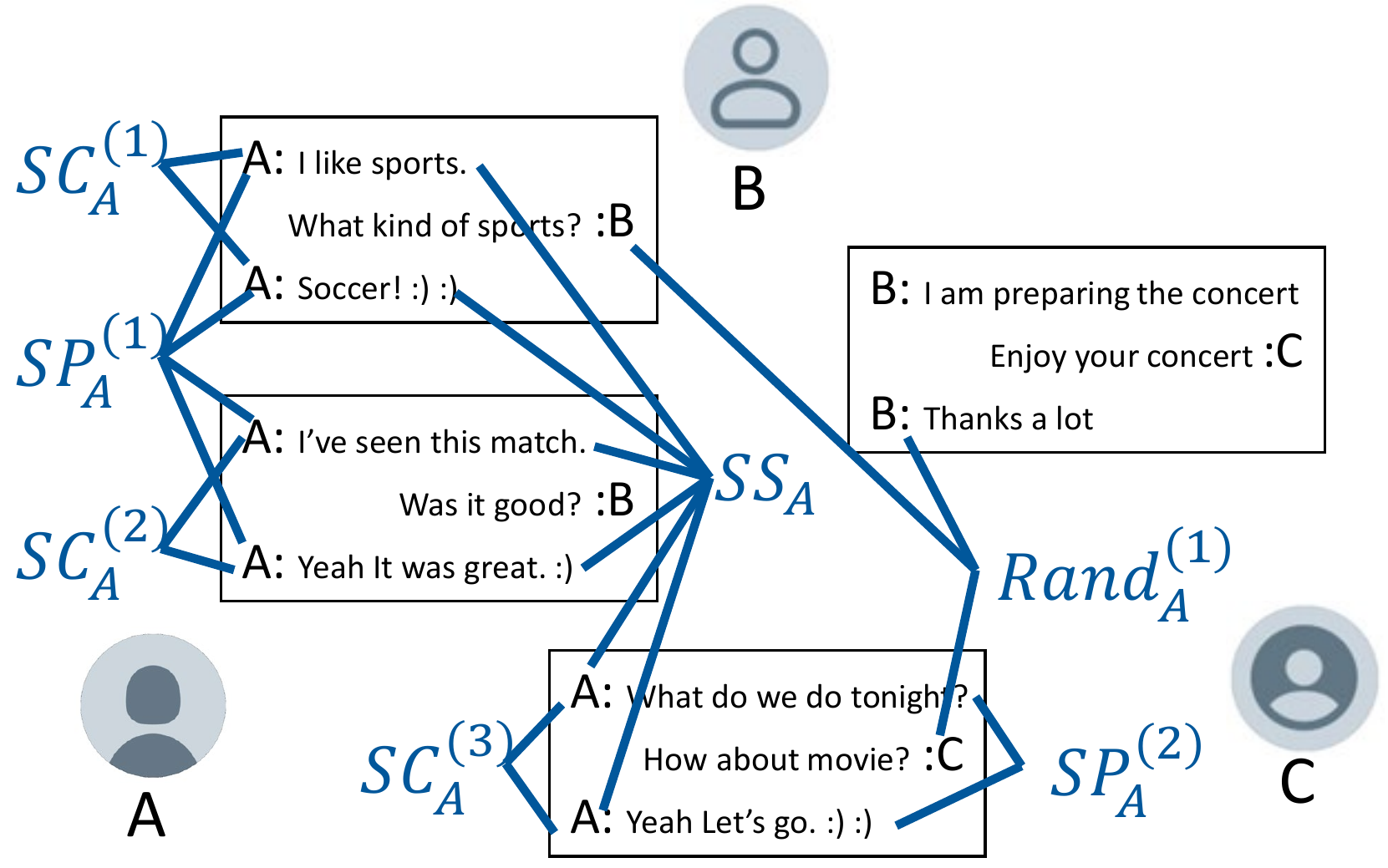}
    \caption{
    Example of utterance sets for speaker \textit{A}.
    $SC$ stands for `same conversation', $SP$ for `same partner',
    $SS$ for `same speaker', and $Rand$ for `random'.
    }
	\label{fig:metric_motivation}
\end{figure}

\begin{table}[t]
    \centering
    \small
    \begin{tabular}{rrrr}
        \toprule
        \multicolumn{1}{c}{$SC$}   & \multicolumn{1}{c}{$SP$}   & \multicolumn{1}{c}{$SS$}   & \multicolumn{1}{c}{$Rand$}   \\
        \midrule
        .922$\pm$1e-4 & .919$\pm$2e-4 & .912$\pm$3e-4 & .898$\pm$2e-3 \\
        \bottomrule
        \end{tabular}
    \caption{Mean similarity among utterances in $SC$, $SP$, $SS$ and $Rand$ sets with a 95\% confidence interval}
    \label{tab:metric_neg_sets_results}
\end{table}

We are motivated by the assumption that there is varying degree of similarity among utterances in a corpus of conversations containing many speakers and conversations.
\begin{enumerate}
\item If we pick a set of random utterances from the corpus, they will not be very similar.
\item If we pick a set of utterances from a single speaker conversing with multiple partners, those utterances will be more similar than the random utterances in 1.
    \item If we pick a set of utterances from conversations between a single dyad, even if the conversations are far apart in time, those utterances would be more similar than those in 2.
    \item If we pick a set of utterances in a single conversation session, they are the most similar, even more so than those in 3.
\end{enumerate}

To test these assumptions, we first categorize one speaker \textit{A}'s utterances into four types of sets corresponding to the assumptions above.
\begin{itemize}
\item Random ($Rand_A$): Random utterances from  speakers who are not \textit{A}
\item Same Speaker ($SS_A$): Speaker \textit{A}'s utterances
\item Same Partner ($SP_A$): \textit{A}'s utterances in conversations with the same partner \textit{B}
    \item Same Conversation ($SC_A$): \textit{A}'s utterances in a single conversation
\end{itemize}
Figure \ref{fig:metric_motivation} shows one example of the sets.
We make three $SC_A$ sets because \textit{A} participates in three conversations.
We make two $SP_A$ sets because \textit{A} has conversations with \textit{B} and \textit{C}.
$SS_A$ is all utterances from \textit{A} so we create one set of utterances for \textit{A}.
Finally, $Rand_A$ is random utterances from non-\textit{A}'s utterances.
We create five sets for each speaker.

From these sets, we compute the similarity among utterances in a set.
First, we convert an utterance into a vector by averaging the words in the utterance with GloVe Twitter 200d \cite{pennington2014glove}.
And we compute the similarity of the vectors by Frobenius norm.
Finally, we calculate the mean similarity of each set with a 95\% confidence interval.
Table \ref{tab:metric_neg_sets_results} shows the results.
$Rand$ has the lowest similarity mean value, so it supports the first assumption.
$SS$ has higher similarity mean value than $Rand$.
It supports the second assumption.
The mean similarity value of $SP$ is higher than $SS$.
It supports the third assumption.
Finally, $SC$ has the highest mean similarity value.
It also supports the last assumption.
From the observations, we assume that utterances are clustered by the speakers and addressees.

\subsection{SSREM}
\label{sec:SSREM_structure}

SSREM evaluates a generated response $\hat{\mathbf{r}}$ from a context $\mathbf{c}$ and a ground truth response $\mathbf{r}$.
The output of SSREM is as follows:
\begin{align}
    SSREM(\mathbf{c}, \mathbf{r}, \hat{\mathbf{r}}) = h(f(\mathbf{c}, \hat{\mathbf{r}}), g(\hat{\mathbf{r}}, \mathbf{r}))
    \label{eqn:ssrem_score}
\end{align}
where $f(\mathbf{c}, \hat{\mathbf{r}}) = tanh(V(\mathbf{c})^T \mathbf{M} V(\hat{\mathbf{r}}))$ is a parametrized function to measure the similarity between the context $\mathbf{c}$ and the generated response $\hat{\mathbf{r}}$.
$V$ is a function to convert a sequence of words to a vector.
$\mathbf{M}$ is a matrix that weights of the similarity between two vectors.
It is the parameter of the $f$ function.
$g(\mathbf{r}, \hat{\mathbf{r}})$ is another function to measure the ground-truth response and the generated one.
$h$ is a function to mix the values of $f$ and $g$ functions.
To normalize each output of the $f$ and $g$ functions, we adopt linear scaling to unit range \cite{aksoy2001feature} which rescale the value $x$ as follows:
\begin{align}
    \tilde{x} = \frac{x - l}{u - l}
\end{align}
where $u$ is an maximum and $l$ is minimum of $x$.

SSREM is similar to RUBER, which computes the similarities among $\mathbf{c}$, $\mathbf{r}$ and $\hat{\mathbf{r}}$ separately and merge it at the end.
However, SSREM uses speaker sensitive samples, whereas RUBER takes one positive sample and one negative sample.

\subsection{Training with Speaker Sensitive Samples}
\label{sec:SSREM_training}

SSREM has a parametrized function $f$ that takes context $\mathbf{c}$ and a generated response $\hat{\mathbf{r}}$.
To train the $f$ function, we define a classification problem to identify the ground truth response $\mathbf{r}$ from a set of candidate responses $R_{cand}$.
The $R_{cand}$ has the ground truth response and some negative samples.
A classifier tries to identify the ground truth response with the negative samples.
Negative samples are usually selected from the uniform distribution.
But we sample the speaker sensitive utterances which described in section \ref{sec:SSREM_motivation} for SSREM.

Formally speaking, let $A$ be the speaker of the ground truth response $\mathbf{r}_A$.
It means it is $A$'s turn to say the response for the context $\mathbf{c}$.
The candidate response set $R_{cand_A}$ is given by 
\begin{align}
    R_{cand_A} = \{ \mathbf{r}_A, \mathbf{sc}_A, \mathbf{sp}_A, \mathbf{ss}_A, \mathbf{rand}_A\}   
\end{align}
where $\mathbf{sc}_A \in SC_A \setminus \mathbf{c}$, $\mathbf{sp}_A \in SP_A \setminus \mathbf{c}$, $\mathbf{ss}_A \in SS_A \setminus \mathbf{c}$ and $\mathbf{rand}_A \in Rand_A$ are the negative samples from speaker sensitive responses.
Then, the probability of a ground truth response $\mathbf{r}_A$ given context $\mathbf{c}$ and $R_{cand_A}$ is as follows:
\begin{align}
    p(\mathbf{r}_A | \mathbf{c}, R_{cand_A}) = \frac{exp(f(\mathbf{c}, \mathbf{r}_A))}{\sum_{\mathbf{r}' \in R_{cand_A}} exp(f(\mathbf{c}, \mathbf{r}'))}
\end{align}
We maximize this probability among all context-ground truth response pair.
So the loss function of the classification problem is
\begin{align}
    - \sum_{\mathbf{c}} \log \frac{exp(f(\mathbf{c}, \mathbf{r}_A))}{\sum_{\mathbf{r}' \in R_{cand_A}} exp(f(\mathbf{c}, \mathbf{r}'))}
\end{align}

This approach is similar to learning the sentence representations \cite{logeswaran2018an}, but we use the speaker sensitive negative samples.
It is also similar to Noise Contrastive Estimation (NCE) \cite{pmlr-v9-gutmann10a,Mnih:2012:FSA:3042573.3042630}.
But we set the noise distribution to speaker sensitive distribution and only take the data sample term in the objective function of the NCE.

Selecting negative samples is important for learning.
When we choose the noise distribution, it would be close to the data distribution, because otherwise, the classification problem might be too easy to learn the data \cite{pmlr-v9-gutmann10a}.
\newcite{Mnih:2012:FSA:3042573.3042630} shows that using samples from the unigram distribution outperforms using samples from a naive uniform distribution for learning a neural probabilistic language model.
Likewise, we create negative samples from the speaker sensitive utterances.
$\mathbf{sc}_A$ is more similar to the $\mathbf{r}_A$ than any other negative samples.
We show the patterns by empirical observations in section \ref{sec:SSREM_motivation} and experimental results in section \ref{sec:exp2-results}.
These speaker sensitive samples make the classification problem harder and lead to learning the function $f$ better than using the naive uniform distributed random samples.

To train SSREM, we need a conversation corpus that has many conversations from one speaker.
We choose the Twitter conversation corpus \cite{bak-oh-2019-variational}
 as it has 770K conversations with 27K Twitter users.
We split the data as 80/10/10 for training/validation/test.

\section{Annotating Human Scores}
\label{sec:exp1_metric_annotation}

To measure the correlation SSREM with human judgments, we first gather human judgments of responses given a conversation context.
We use Amazon Mechanical Turk (MTurk) to annotate the scores of the responses.
We select 300 conversations from a dataset of Twitter conversations.
And we generate responses for annotation using three conversation models and the ground truth response for each conversation.
\begin{itemize}
    \item Retrieval model \cite{pandey-etal-2018-exemplar}: A BM25 retrieval model \cite{robertson2009probabilistic} that uses TF-IDF vector space.
    \item VHCR \cite{VHCR:2018:NAACL}: A variational autoencoder model that has a global variable for a conversation.
    \item VHUCM \cite{bak-oh-2019-variational}: A variational autoencoder model that considers the speakers of a conversation.
\end{itemize}
Then we ask two questions to the MTurkers.
(1) How appropriate is the response overall?
(2) How on-topic is the response?
These questions are used in \cite{lowe-etal-2017-towards}.
The authors show that these questions have high inter-annotator agreement among workers.
They suggest using the first question to annotate the human score, and so we follow the suggestion.
But we ask the second question to workers to filter out workers who submit random answers.
Each worker answers these questions on a five-point Likert scale.

\begin{table}[t]
    \centering
    \begin{tabular}{lrrrrr}
    \toprule
    Human Score    & 1   & 2   & 3   & 4   & 5   \\
    \midrule
    Twitter & 211 & 258 & 342 & 278 & 71\\
    Movie & 279 & 267 & 311 & 217 & 126\\
    \bottomrule
    \end{tabular}
    \caption{Basic statistics of human scores of the responses on Twitter conversation and Movie scripts}
    \label{tab:metric_basic_stats}
\end{table}

We annotate 1,200 responses in total.
One worker answers ten conversations, four responses per conversation for a total of 40 responses.
Each response is tagged by five workers for a total of 287 workers of which we retain the responses from 150 workers who passed all the tests.
We tag the most selected score as the human score for each response.
The inter-annotator Fleiss' kappa \cite{fleiss1971measuring} is $\kappa = 0.61$ which is consistent with the results in \cite{lowe-etal-2017-towards}.
Table \ref{tab:metric_basic_stats} shows the basic statistics of the annotations.

\section{Experiment 1 - Comparing with Human Scores}
\label{sec:exp1_results}

This section describes the experiment that looks at the correlation between the model scores and the human scores for given contexts and responses.

\subsection{Experiment Setup}
We use a Twitter conversation corpus \cite{bak-oh-2019-variational} to train and validate SSREM and other baseline models.
For the test, we remove the ground truth responses in human-annotated corpus since it always produces the maximum score on BLEU and ROUGE.

We compare SSREM with the following response evaluation methods:
\begin{itemize}
    \item BLEU \cite{P02-1040}: We compute the sentence-level BLEU score with the smoothing seven technique \cite{W14-3346}.
    \item ROUGE \cite{W04-1013}: We compute the F score of ROUGE-L.
    \item EMB \cite{D16-1230}: We compute the average cosine similarity between ground truth response and test response in a word embedding\footnote{We experimented with the greedy and extreme embedding for comparison, but these methods were not better than the average embedding.}.
    We use pre-trained Google news word embedding \cite{NIPS2013_5021} to avoid the dependency between the training data and embedding.
    \item RUBER \cite{tao2018ruber}: We train with a random negative sample to train unreferenced metric in RUBER. And we use arithmetic averaging to hybrid the referenced and unreferenced metrics.
    \item RSREM: We use the same structure of SSREM, but train with uniformly random negative samples, not speaker sensitive samples.
\end{itemize}

We choose functions in SSREM for the experiment.
For $V$ function, We use the word averaging technique that averages the vectors of words in the sequence.
We can use advanced methods such as RNN or sentence embeddings \cite{reimers-gurevych-2019-sentence}.
But for the fair comparisons with RUBER, we select a similar approach.
We use GloVe Twitter 200d word embedding \cite{pennington2014glove}.
For $g$ function, we use sentence mover`s similarity that is the state of the art evaluating reference-candidate pair of sentences by using word and sentence embeddings \cite{clark-etal-2019-sentence}.
To avoid dependency between the training data and embedding, we use Elmo embedding \cite{Peters:2018}.
For $h$ function, we use arithmetic averaging that shows good results in \cite{tao2018ruber}.

\subsection{Results and Discussion}

\begin{table}[t]
    \centering
    \begin{tabular}{lll}
      \toprule
    \multicolumn{1}{c}{Metric} & \multicolumn{1}{c}{Spearman}            & \multicolumn{1}{c}{Pearson}                   \\
    \midrule
    BLEU   & 0.024 $(0.472)$             & 0.041  $(0.227)$                  \\
    ROUGE  & 0.024  $(0.471)$            & 0.052  $(0.124)$                   \\
    EMB    & 0.006   $(0.861)$           & 0.012   $(0.720)$                \\
    RUBER  & 0.044   $(0.192)$           & 0.046     $(0.177)$              \\
    RSREM  & 0.088  $(<0.01)$            & 0.101   $(<0.01)$                \\
    SSREM & \textbf{0.392} $(<0.001)$             & \textbf{0.376}  $(<0.001)$                 \\
    \bottomrule
    \end{tabular}
    \caption{Correlation between human and model scores. 
    We compute Spearman and Pearson correlation coefficients.
    $p$-values are shown in brackets.
    SSREM shows higher correlation with human judgement than the other models.
    }
      \label{tab:correlations_human_models_tc}
    \end{table}

\begin{figure*}[t]
	\centering
	\begin{subfigure}[b]{0.32\textwidth}
		\includegraphics[width=\textwidth]{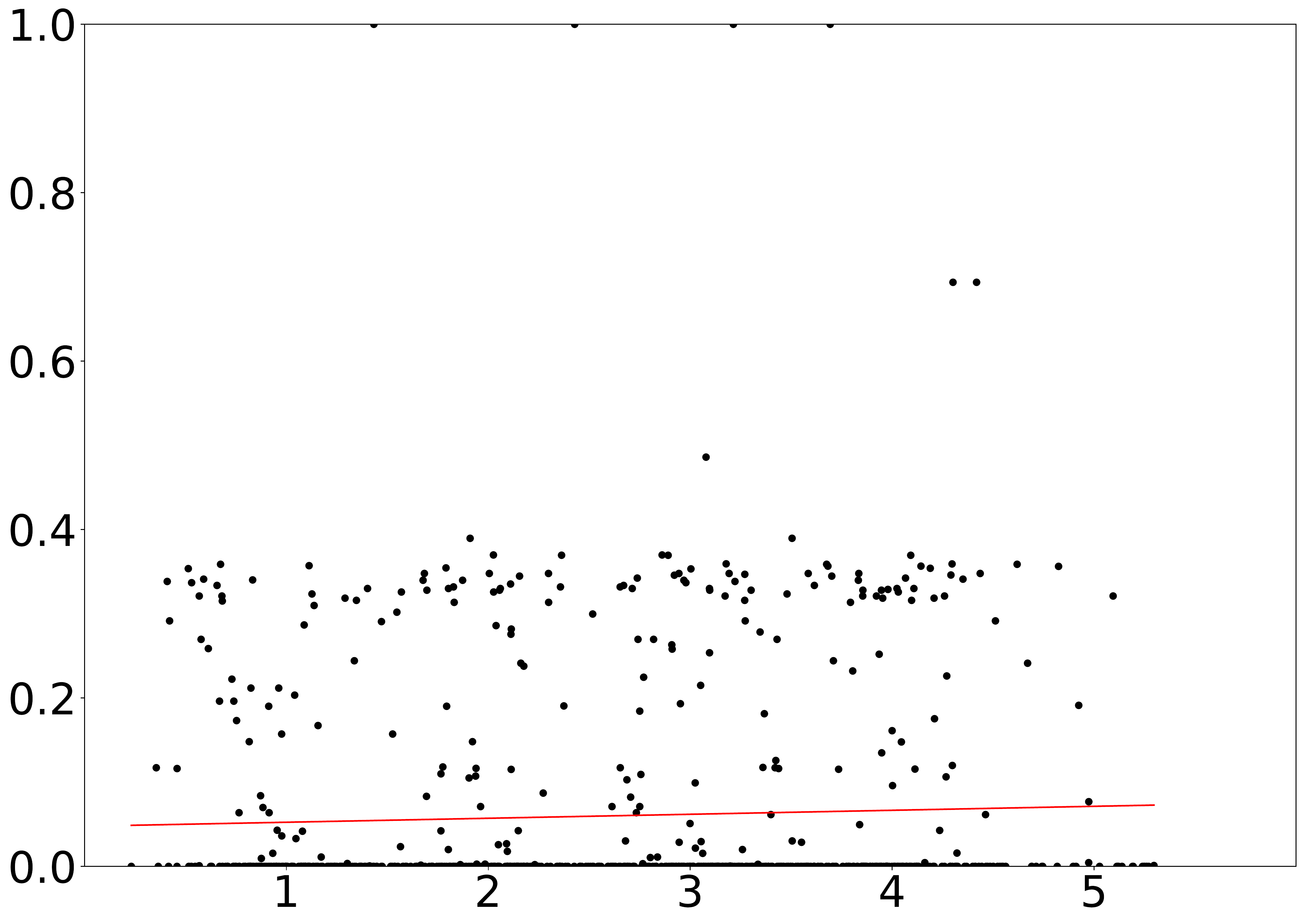}
		\caption{BLEU, coeff: $0.005$}
	\end{subfigure}
	\begin{subfigure}[b]{0.32\textwidth}
		\includegraphics[width=\textwidth]{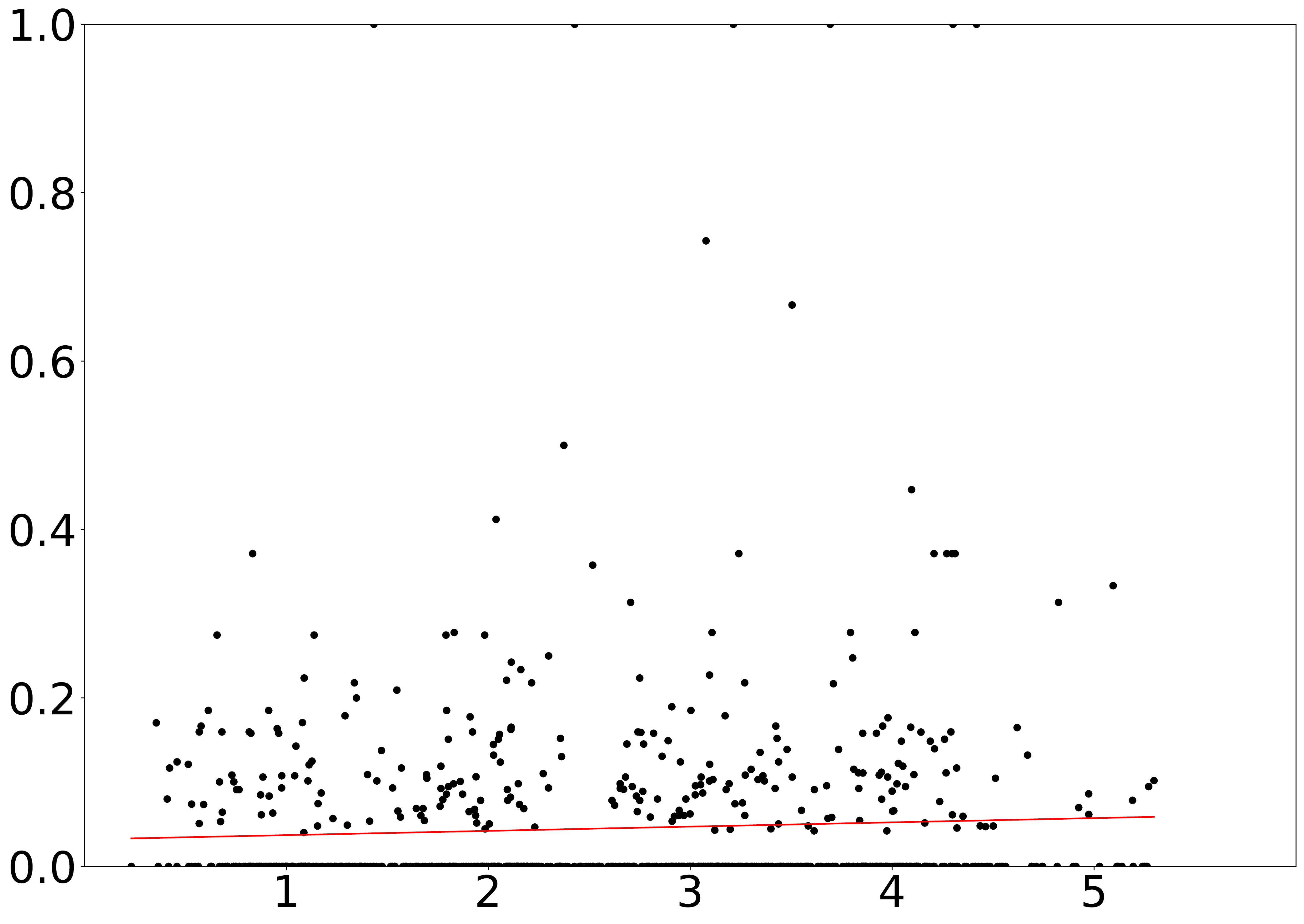}
		\caption{ROUGE, coeff: $0.005$}
	\end{subfigure}
    \begin{subfigure}[b]{0.32\textwidth}
		\includegraphics[width=\textwidth]{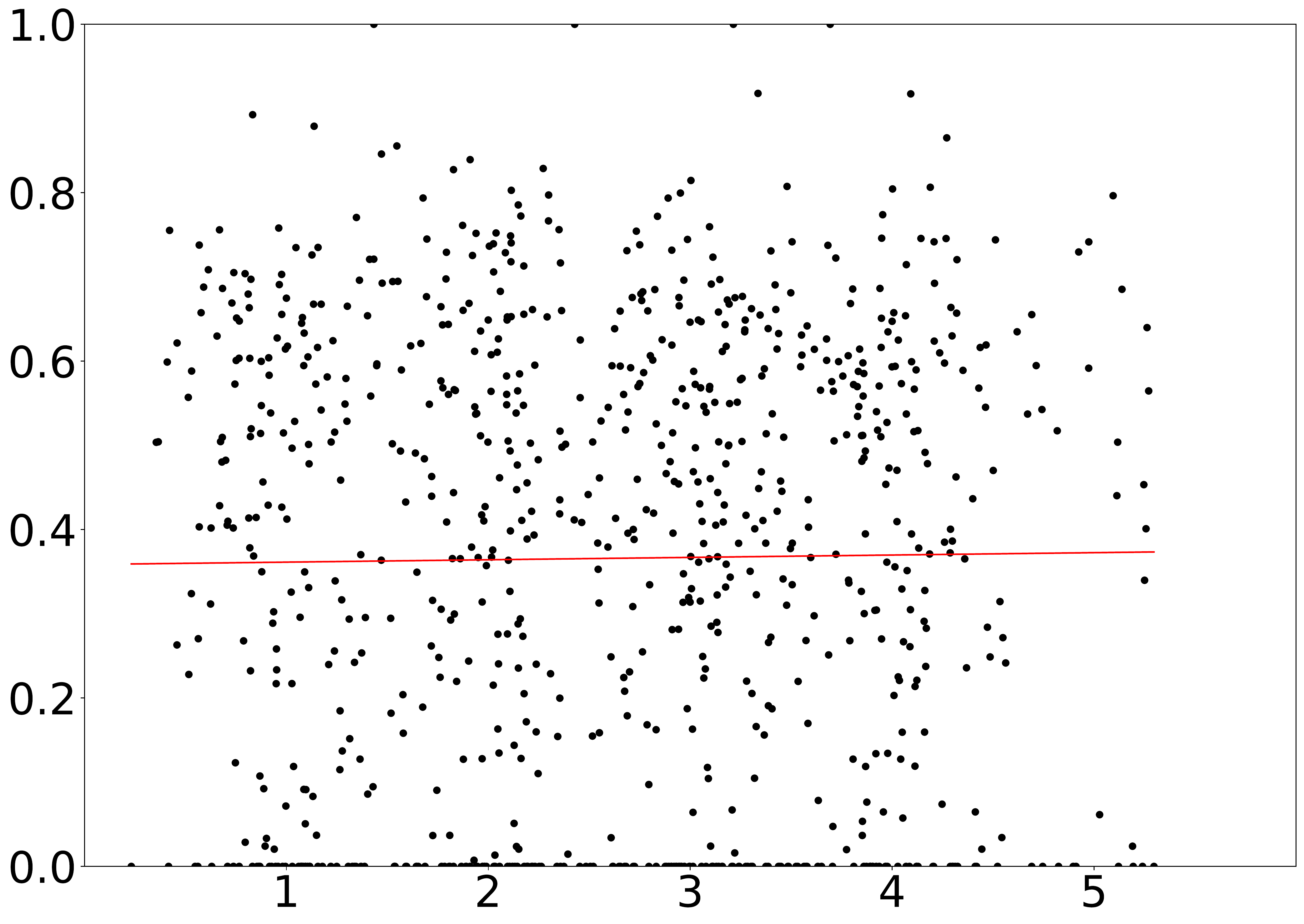}
		\caption{EMB, coeff: $0.003$}
    \end{subfigure}
    \begin{subfigure}[b]{0.32\textwidth}
		\includegraphics[width=\textwidth]{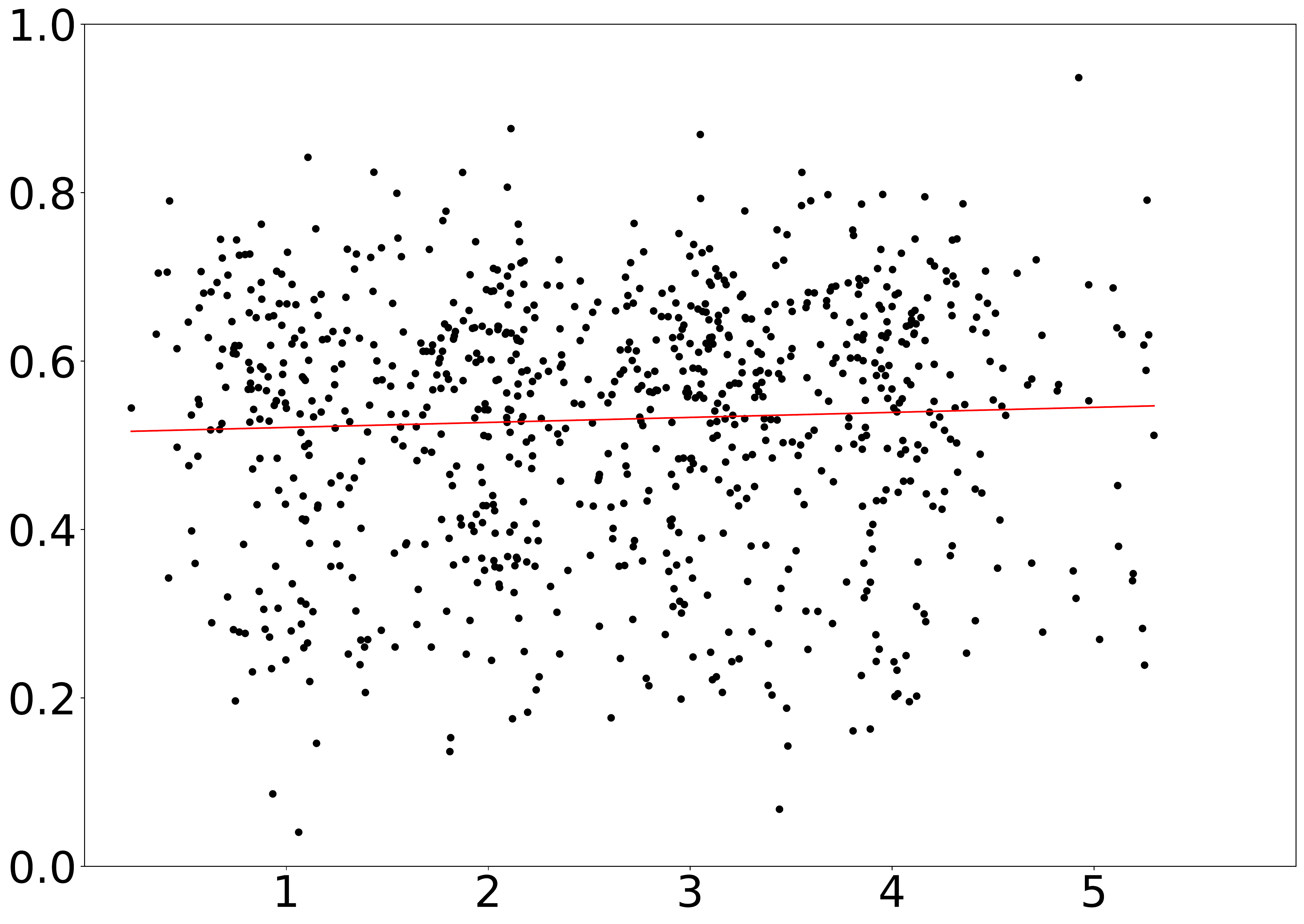}
		\caption{RUBER, coeff: $0.006$}
    \end{subfigure}
    \begin{subfigure}[b]{0.32\textwidth}
		\includegraphics[width=\textwidth]{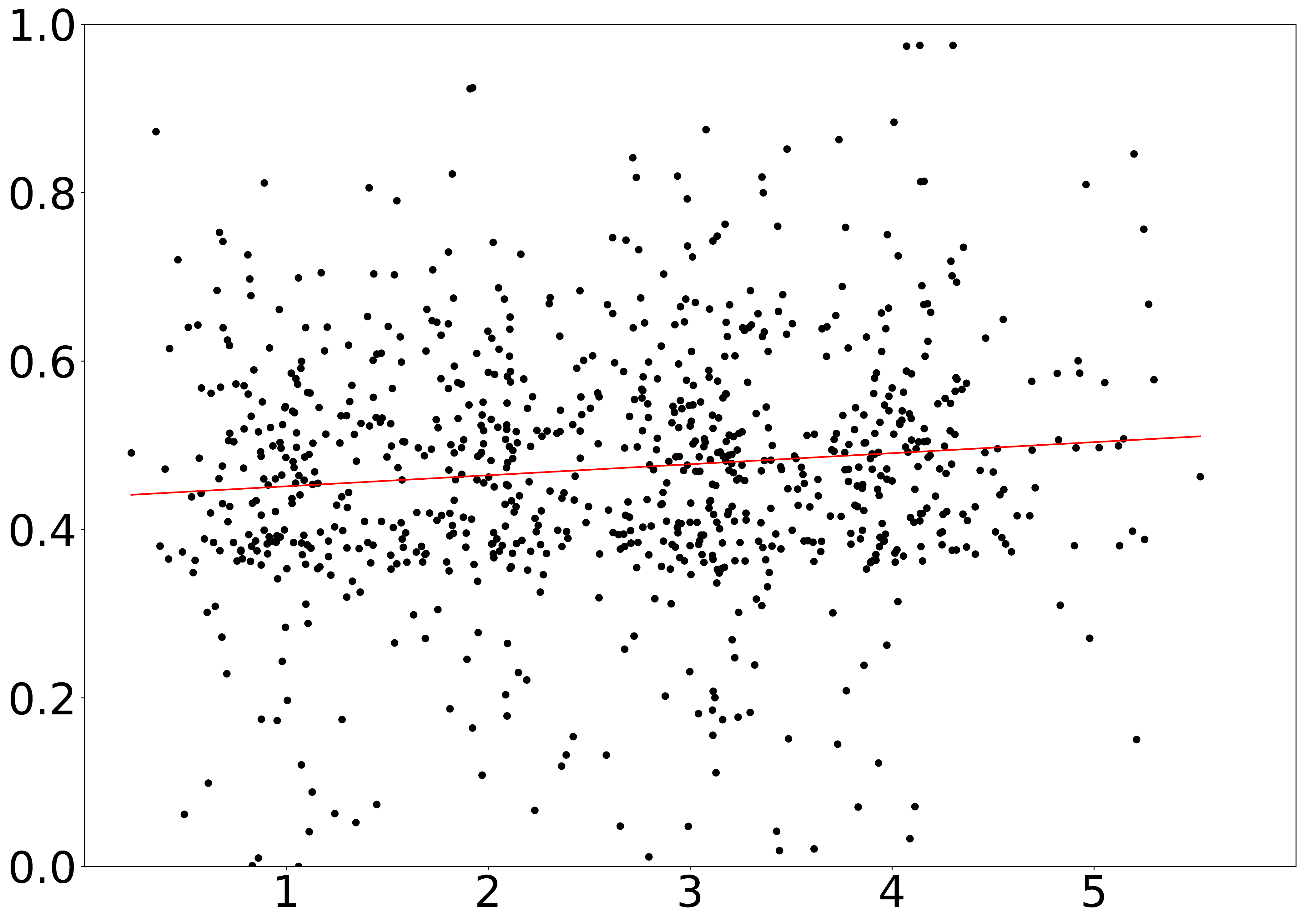}
		\caption{RSREM, coeff: $0.013$}
    \end{subfigure}
    \begin{subfigure}[b]{0.32\textwidth}
		\includegraphics[width=\textwidth]{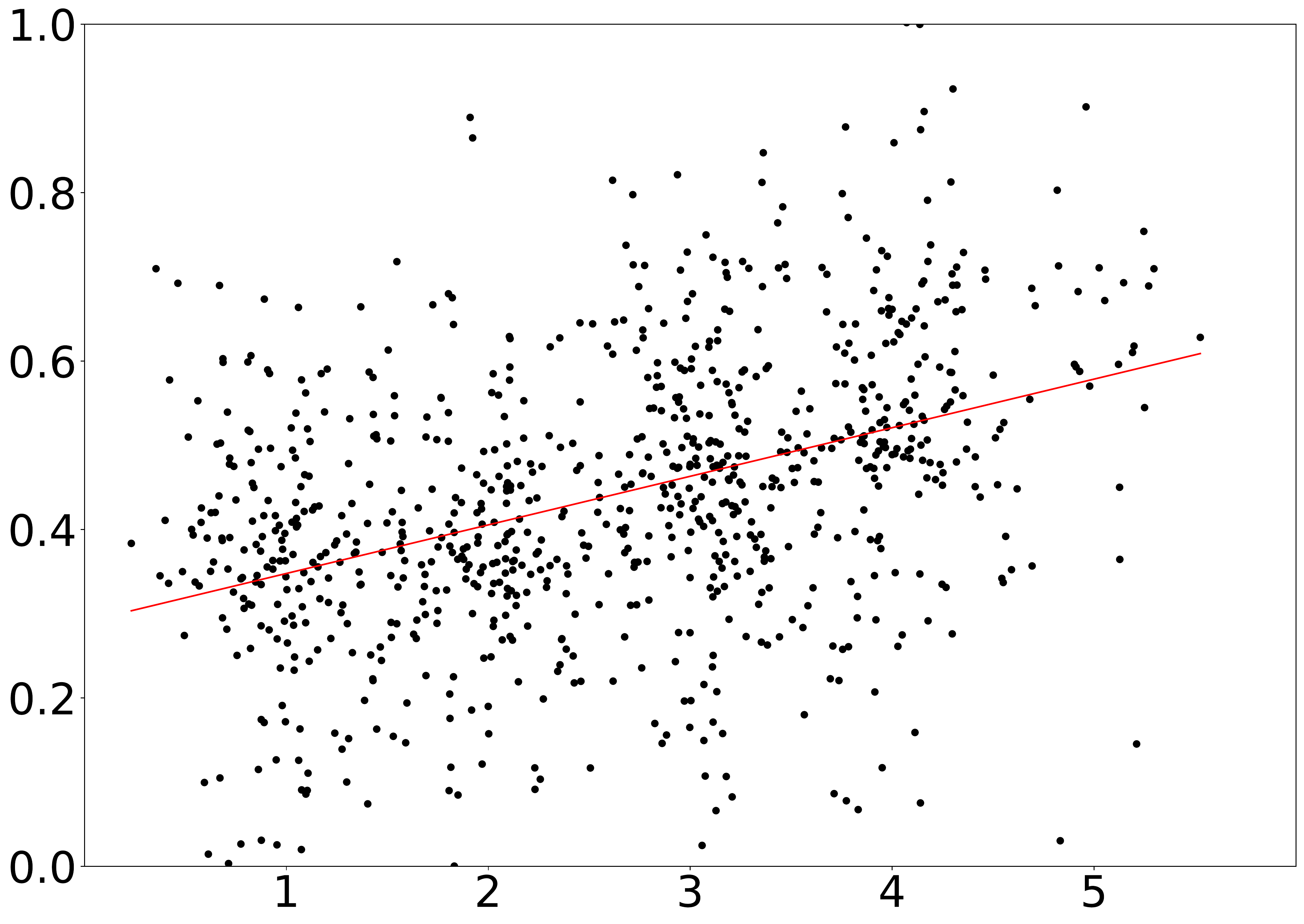}
		\caption{SSREM, coeff: $0.058$}
    \end{subfigure}
    \caption{Scatter plots that show model scores against human scores.
    We add Gaussian noise drawn from $N(0, 0.3)$ to the human scores to better visualize the density of points \cite{lowe-etal-2017-towards}.
    The red line is a linear regression line, and the coeff is the coefficient of the line.
    SSREM shows a higher positive correlation with human judgment than the other models.}
	\label{fig:metric_human_relationship_scatter_tc}
\end{figure*}

Table \ref{tab:correlations_human_models_tc} shows the Spearman and Pearson correlations between human scores and models scores.
First, BLEU, ROUGE, and EMB are not correlated with human scores.
It means evaluating responses with ground truth only is not useful.
These results are the same in previous research \cite{D16-1230,lowe-etal-2017-towards,tao2018ruber}.
RUBER shows a higher correlation with human scores than other baselines but has a high $p$-value that means low statistically significant.
RSREM performs better than RUBER and other baselines.
It shows using multiple negative samples improves the performance of learning the model.
Finally, SSREM outperforms all other methods for two correlations with low $p$-values.
It shows the effectiveness of using speaker sensitive negative samples.

Figure \ref{fig:metric_human_relationship_scatter_tc} shows scatterplots of the human and model scores.
A dot is one response, and a red line is a linear regression line.
The x-axis is the human score, and the y-axis is each automatic evaluation metric.
To visualize the dots better, we adopt the technique from \cite{lowe-etal-2017-towards} that adds random number ($N(0, 0.3)$) to x-axis value.
But, we train the linear regression with original scores.
First, BLEU and ROUGE have many zero values since there are few overlapped words between the generated response and the ground-truth response.
The dots in EMB that uses word embedding to overcome the limitation are more distributed.
But there are few relationships with human scores, and the linear regression coefficient is flattened.
RUBER is better than BLEU, ROUGE, and EMB.
RSREM that uses more negative samples shows better than RUBER.
Finally, SSREM shows a higher positive correlation with human scores than other baselines.

\section{Experiment 2 - Identifying True and False Responses}
\label{sec:exp2}
The second experiment presents the performance of $f$ function in SSREM by comparing it with baselines.
RUBER, RSREM, and SSREM compute the score from the context of the conversation and generated responses.
To investigate the performance of the score, we set up the task that identifies the true and false responses for a given context.
The true responses are ground-truth responses, and false ones are four negative samples that are described in section \ref{sec:SSREM_training}.

\subsection{Experiment Setup}
% We compare two baselines with our method.
% \textit{PosOnly} is using positive example only in the likelihood.
% \textit{NegSamUniform} is using positive example and negative samples that are randomly selected.
% \textit{NegSamSpeaker} is our method.
% It uses positive example and negative samples that are utterances by the speaker who says the positive exmaple.

The data for this experiment is the test data of the Twitter conversation corpus.
We extract contexts, true and false responses from the data.
The true response is the ground-truth response ($GT$).
And the false responses are four types that are described in section \ref{sec:SSREM_training} ($SC$, $SP$, $SS$, $Rand$).

We compare SSREM with RUBER and RSREM that compute the similarity between a context and a response.
We take the unreferenced metric score in RUBER.
And we take the output of the $f$ function in RSREM and SSREM.
We use the same trained models in section \ref{sec:exp1_results}.

\subsection{Results and Discussion}
\label{sec:exp2-results}

\begin{figure}[t]
    \centering
		\includegraphics[width=0.45\textwidth]{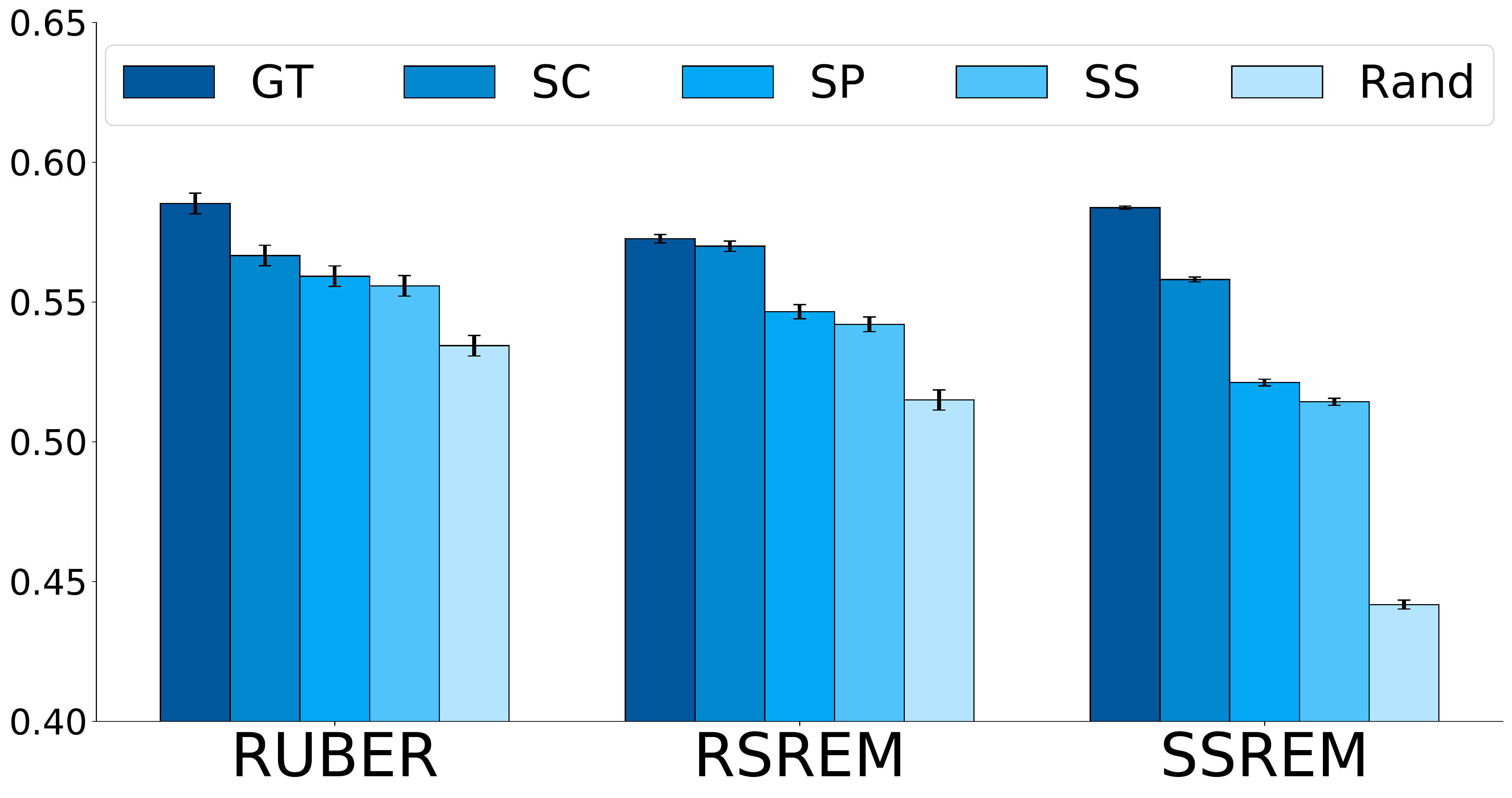}
    \caption{
    Difference of scores on various responses in Twitter conversation corpus.
    The range of the vertical error bar is a 95\% confidence interval of the values among the responses.
    SSREM outperforms the other models for identifying true and false responses.
    }
	\label{fig:metric_pre_result1}
\end{figure}

Figure \ref{fig:metric_pre_result1} shows the results.
The x-axis is the models, and the y-axis is the output of the unreferenced metric or $f$ function.
All models perform well on distinguishing between $GT$ utterances and $Rand$ utterances.
But RUBER performs poor on identifying $SC$, $SP$, and $SS$.
And RSREM cannot identify false responses from $SC$.
Finally, SSREM outperforms the other two models for identifying all cases.
It also maximizes the difference between $GT$ and $Rand$ than the other two models.
It is another clue for showing the effectiveness of using speaker sensitive negative samples.

One interesting result is that the output scores decrease from $GT$ to $Rand$.
It is the same observation about the differences of speaker sensitive utterances in section \ref{sec:SSREM_motivation}.
And it also means that identifying $GT$ and $SC$ is a harder problem than $GT$ and $Rand$ pair.
It is another evidence for why we use speaker sensitive negative samples, as we discussed in section \ref{sec:SSREM_training}.

$SC$ consists of negative samples that are most difficult for the model to distinguish, so it makes sense to consider only $SC$ negative samples. 
But we include $SP$ and $SS$ for the following two reasons.
First, there are only a limited number of $SC$ utterances because they must all come from the same conversation, whereas we need a pretty large number of negative samples to effectively train the model \cite{Mnih:2012:FSA:3042573.3042630}.
Second, we also sample from $SP$ and $SS$ because they represent different degree of similarity to the context utterances.
$SC$ utterances are from the same conversation, leading to decreased model generalization.

\section{Experiment 3 - Applying New Corpus}
\label{sec:exp3}
In this section, we investigate the applicability of SSREM to a new conversation corpus.
SSREM takes the speaker sensitive samples from Twitter.
But there are many open-domain conversation corpora such as Movie scripts \cite{Danescu-Niculescu-Mizil+Lee:11a}.
\newcite{tao2018ruber} run a similar experiment with RUBER, but they use the similar domain of data, Chinese online forum (Training from Douban and testing on Baidu Tieba).
We choose the Movie scripts corpus because it is written by the script writers whereas Twitter is personal causal online conversations.
We present the performance of SSREM on the new corpus.

\subsection{Experiment Setup}
\label{sec:exp2_exp_setup}

First, we annotate 1,200 responses to the movie dialog corpus.
We use HRED \cite{sordoni2015hierarchical} rather than VHUCM.
The next procedure of annotation is the same when we create human scores for Twitter conversation responses in section \ref{sec:exp1_metric_annotation}.
Two hundred forty-four workers tagged all responses.
But, 94 workers failed the attention check question, so we collect the 150 workers' answers.
The inter-annotator Fleiss' kappa \cite{fleiss1971measuring} for Movie is $\kappa = 0.63$.
It is still consistent with the results in \cite{lowe-etal-2017-towards} and annotated Twitter conversations.
The bottom row in Table \ref{tab:metric_basic_stats} shows the basic statistics of the annotated responses.

We run two experiments, comparing with human scores and identifying true and false responses.
We use the same models in section \ref{sec:exp1_results}.
We use the Twitter conversation corpus to train RUBER, RSREM, and SSREM.
And we test the models on annotated movie dialogs.
Unlike the Twitter conversation corpus, the movie dialogs have a short length of conversations. 
So we choose $SC$ and $Ran$ only to run the second experiment.

\subsection{Results and Discussion}
\label{sec:exp2_exp_results}

\begin{figure*}[t]
	\centering
	\begin{subfigure}[b]{0.32\textwidth}
		\includegraphics[width=\textwidth]{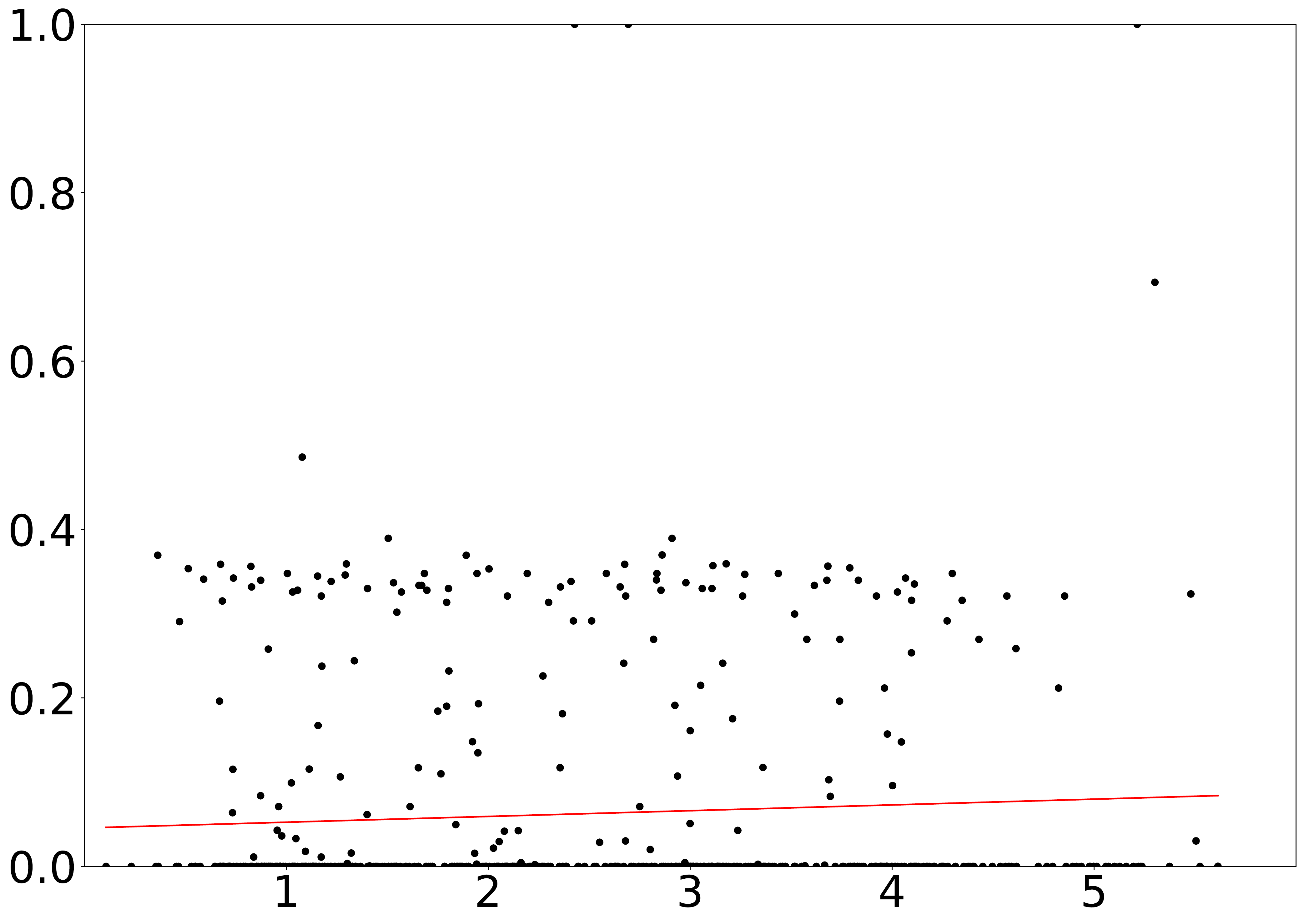}
		\caption{BLEU, coeff: $0.007$}
	\end{subfigure}
	\begin{subfigure}[b]{0.32\textwidth}
		\includegraphics[width=\textwidth]{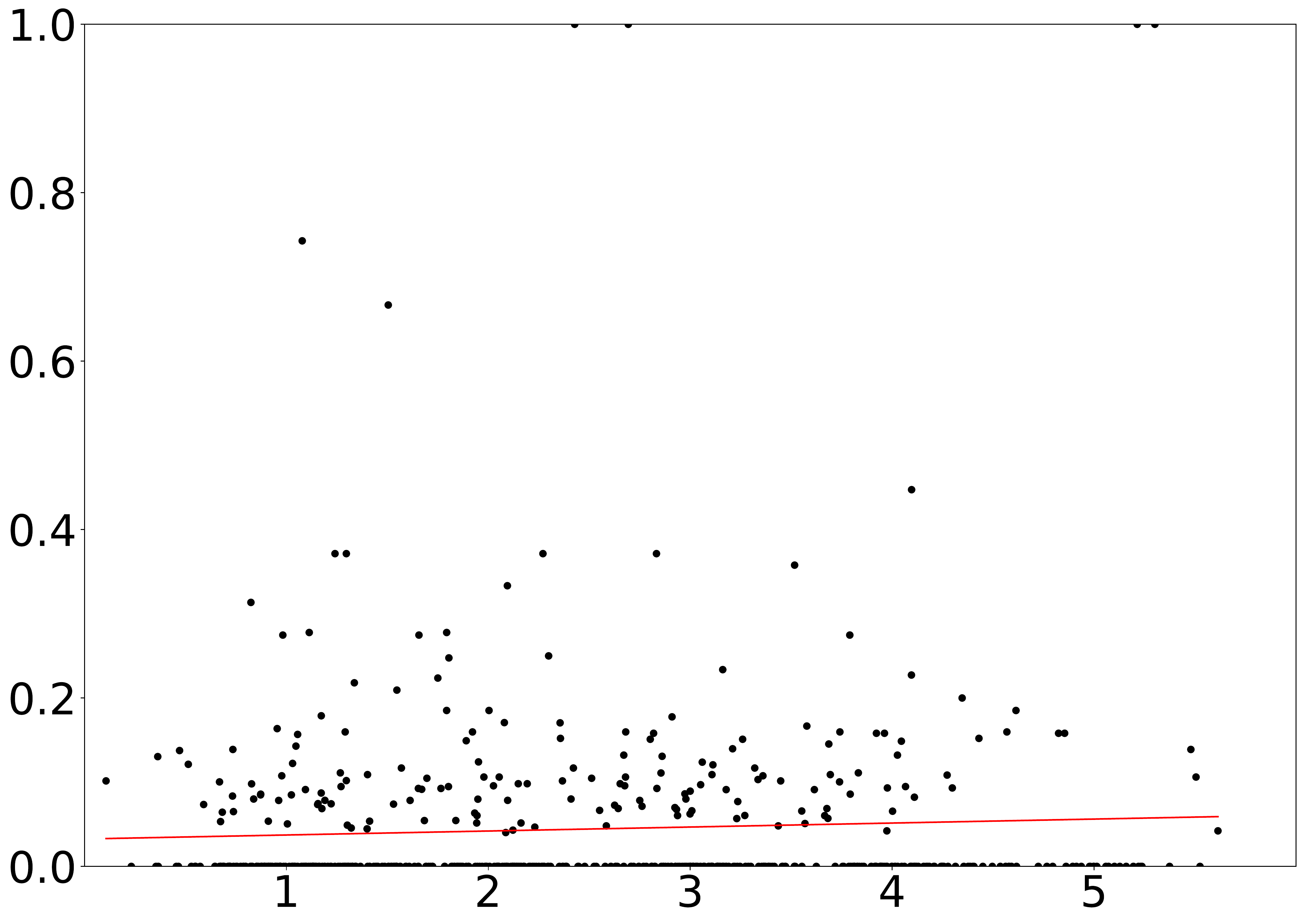}
		\caption{ROUGE, coeff: $0.005$}
	\end{subfigure}
	\begin{subfigure}[b]{0.32\textwidth}
		\includegraphics[width=\textwidth]{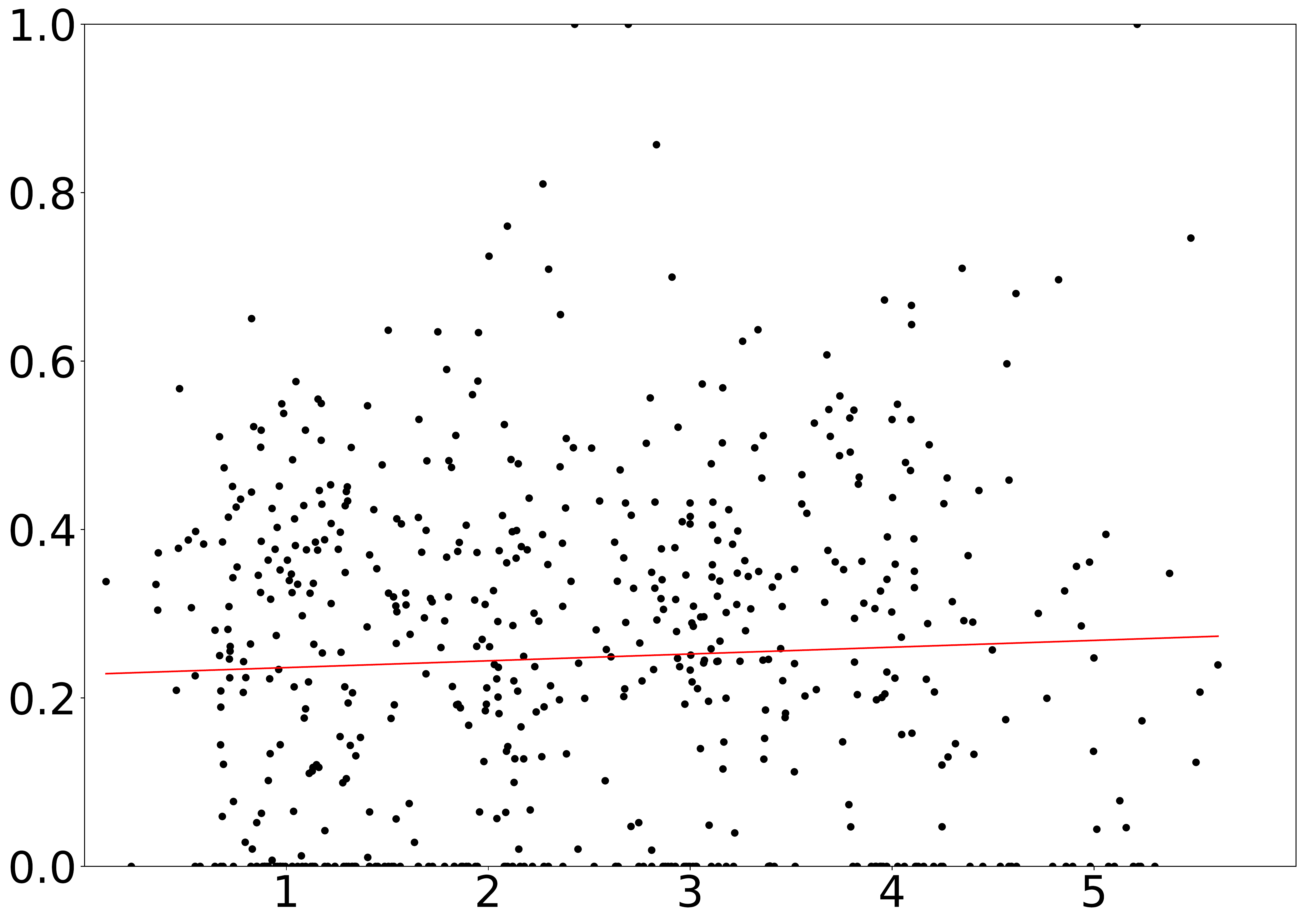}
		\caption{EMB, coeff: $0.002$}
    \end{subfigure}
    \begin{subfigure}[b]{0.32\textwidth}
		\includegraphics[width=\textwidth]{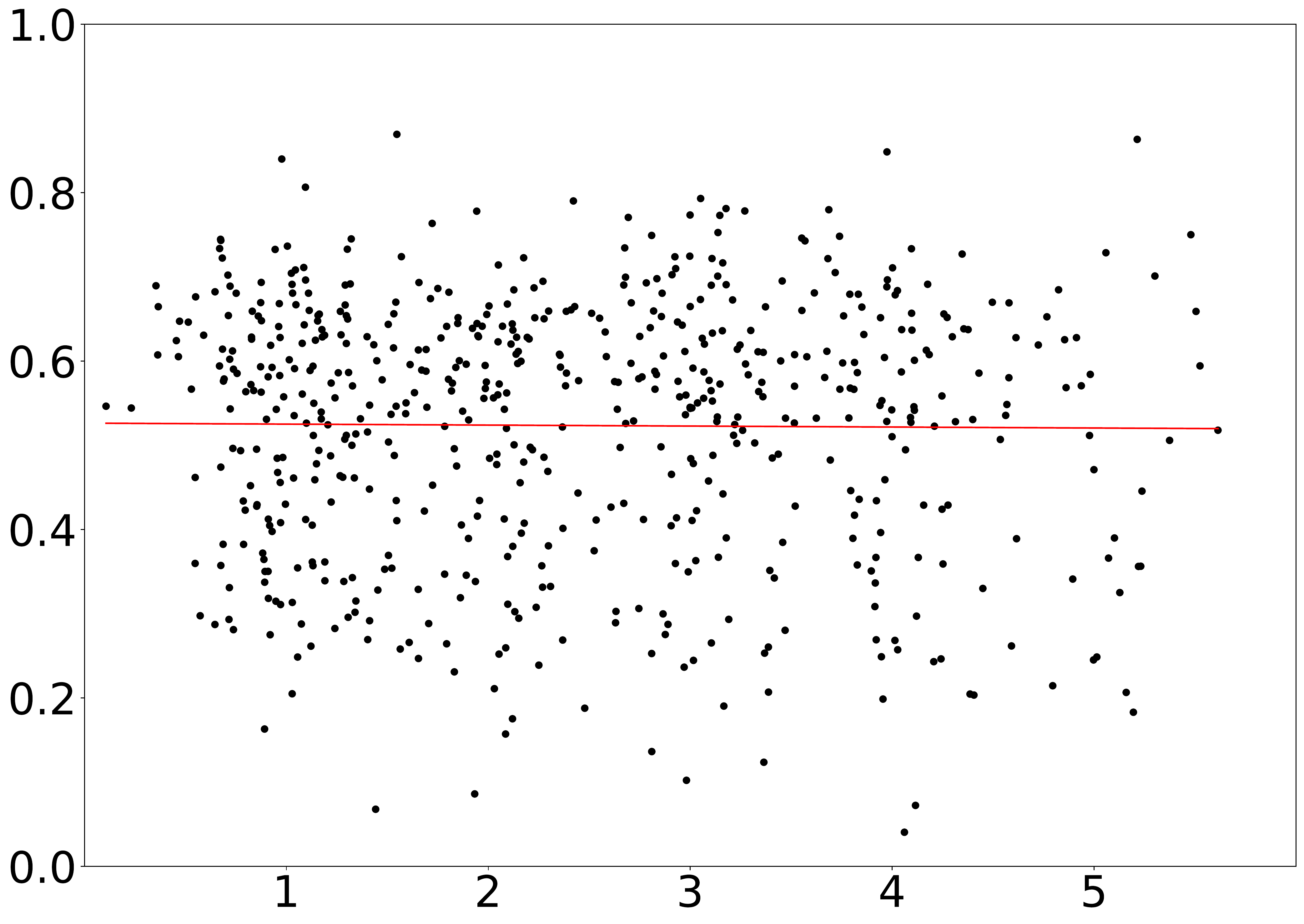}
		\caption{RUBER, coeff: $-0.001$}
    \end{subfigure}
    \begin{subfigure}[b]{0.32\textwidth}
		\includegraphics[width=\textwidth]{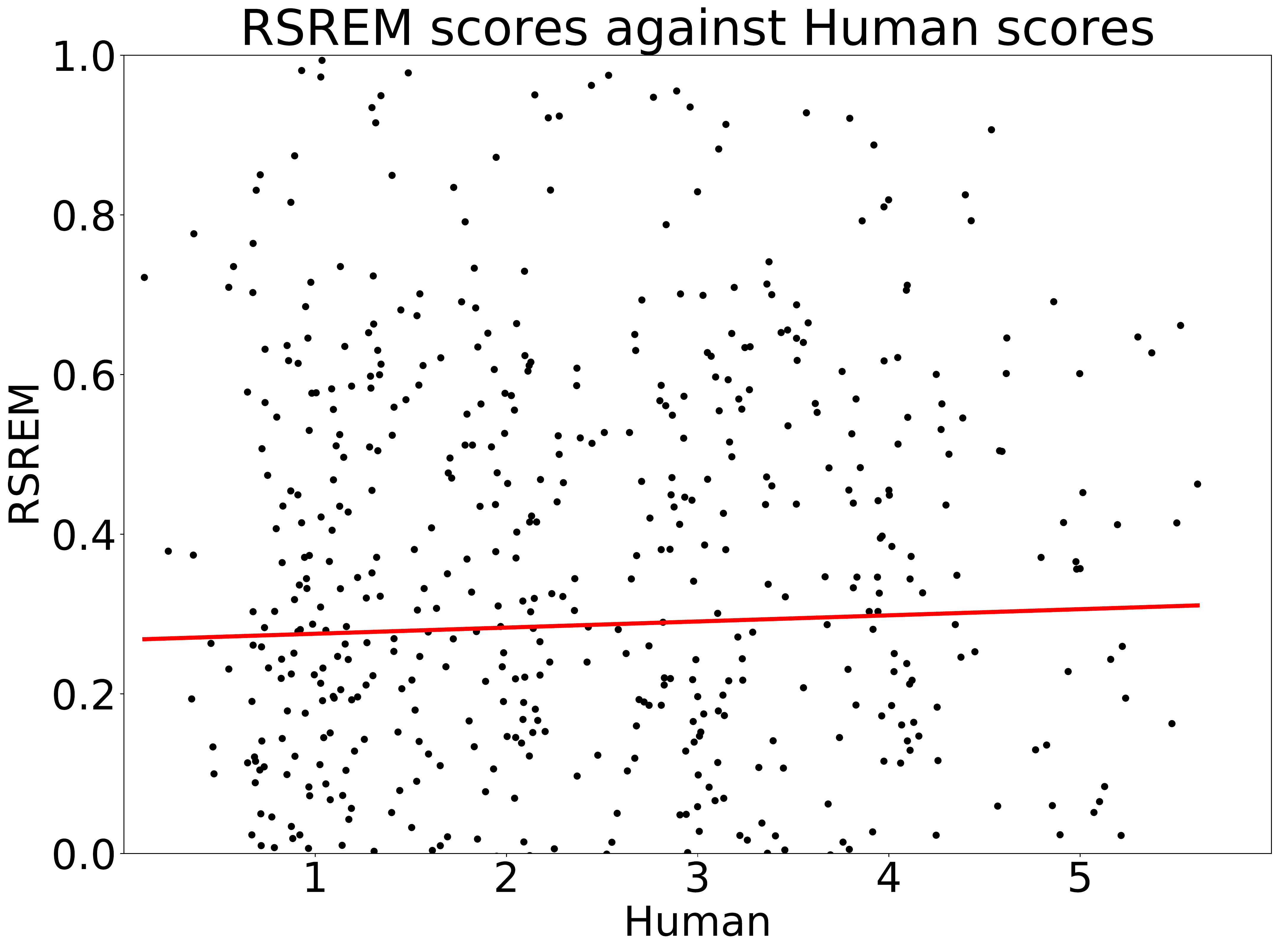}
		\caption{RSREM, coeff: $0.008$}
    \end{subfigure}
    \begin{subfigure}[b]{0.32\textwidth}
		\includegraphics[width=\textwidth]{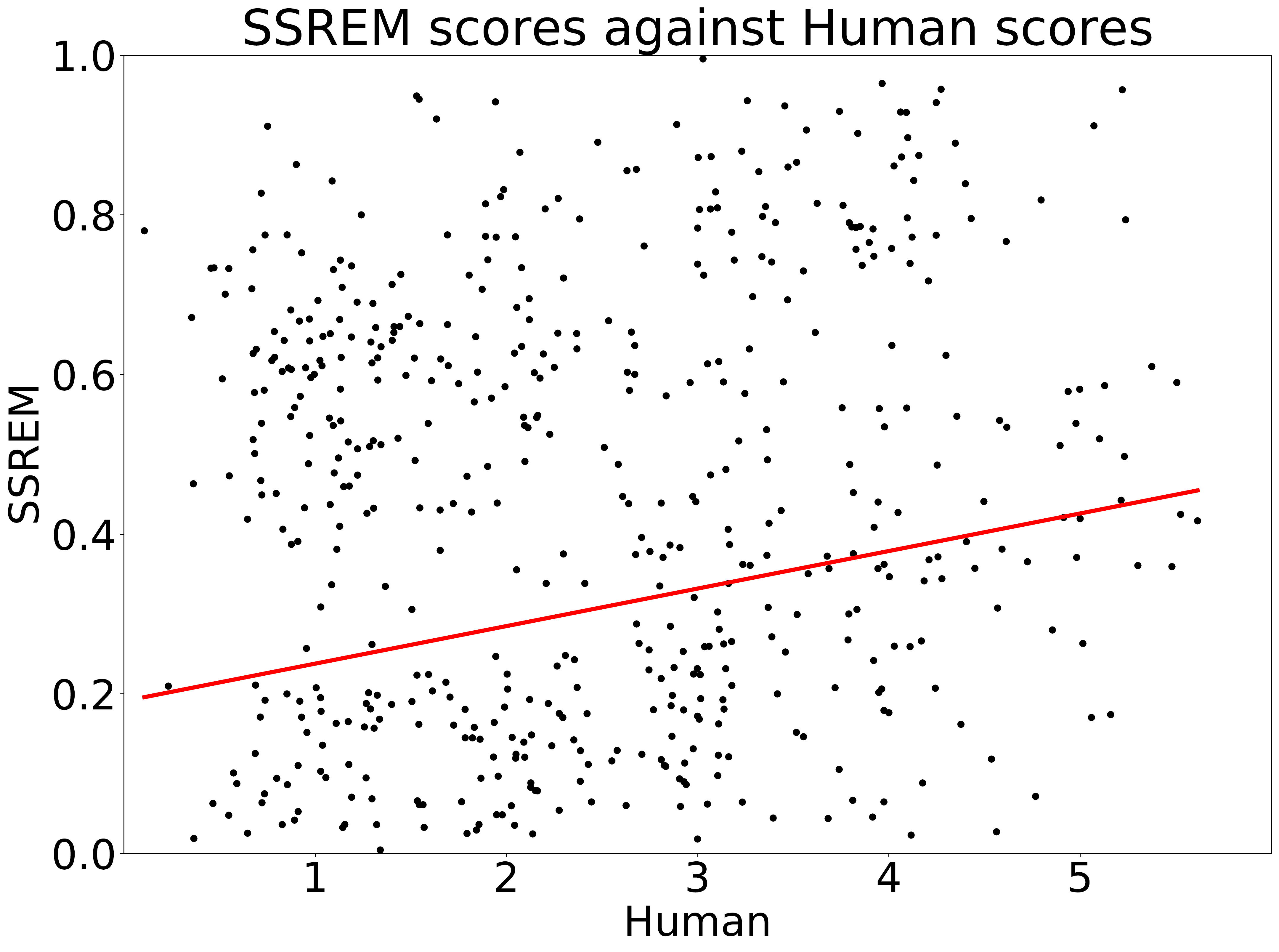}
		\caption{SSREM, coeff: $0.047$}
    \end{subfigure}
    \caption{Scatter plot showing model against human scores with Movie corpus.
    We add Gaussian noise drawn from $N(0, 0.3)$ to the human scores to better visualize the density of points which is similar to \cite{lowe-etal-2017-towards}.}
	\label{fig:metric_human_relationship_scatter_cornell}
\end{figure*}

\begin{table}[t]
    \centering
    \begin{tabular}{lll}
      \toprule
    \multicolumn{1}{c}{Metric} & \multicolumn{1}{c}{Spearman}            & \multicolumn{1}{c}{Pearson}                   \\
    \midrule
    BLEU   & 0.036 $(0.378)$         & 0.063  $(0.124)$       \\
    ROUGE  & 0.041  $(0.322)$         & 0.054  $(0.191)$       \\
    EMB    & 0.022  $(0.586)$        & 0.010  $(0.815)$      \\
    RUBER & 0.004  $(0.920)$        & -0.009   $(0.817)$     \\
    RSREM & 0.009  $(0.817)$         & 0.024  $(0.550)$       \\
    SSSREM & \textbf{0.132}   $(<0.001)$        & \textbf{0.119}  $(<0.005)$      \\
    \bottomrule
    \end{tabular}
    \caption{Correlation between human and model scores with Movie corpus.
    We compute Spearman and Pearson correlation coefficient.
    $p$-values are shown in brackets.
    SSREM shows higher correlation with human judgement than the other models.
    }
      \label{tab:correlations_cornell}
    \end{table}

\begin{figure}[t]
    \centering
		\includegraphics[width=0.45\textwidth]{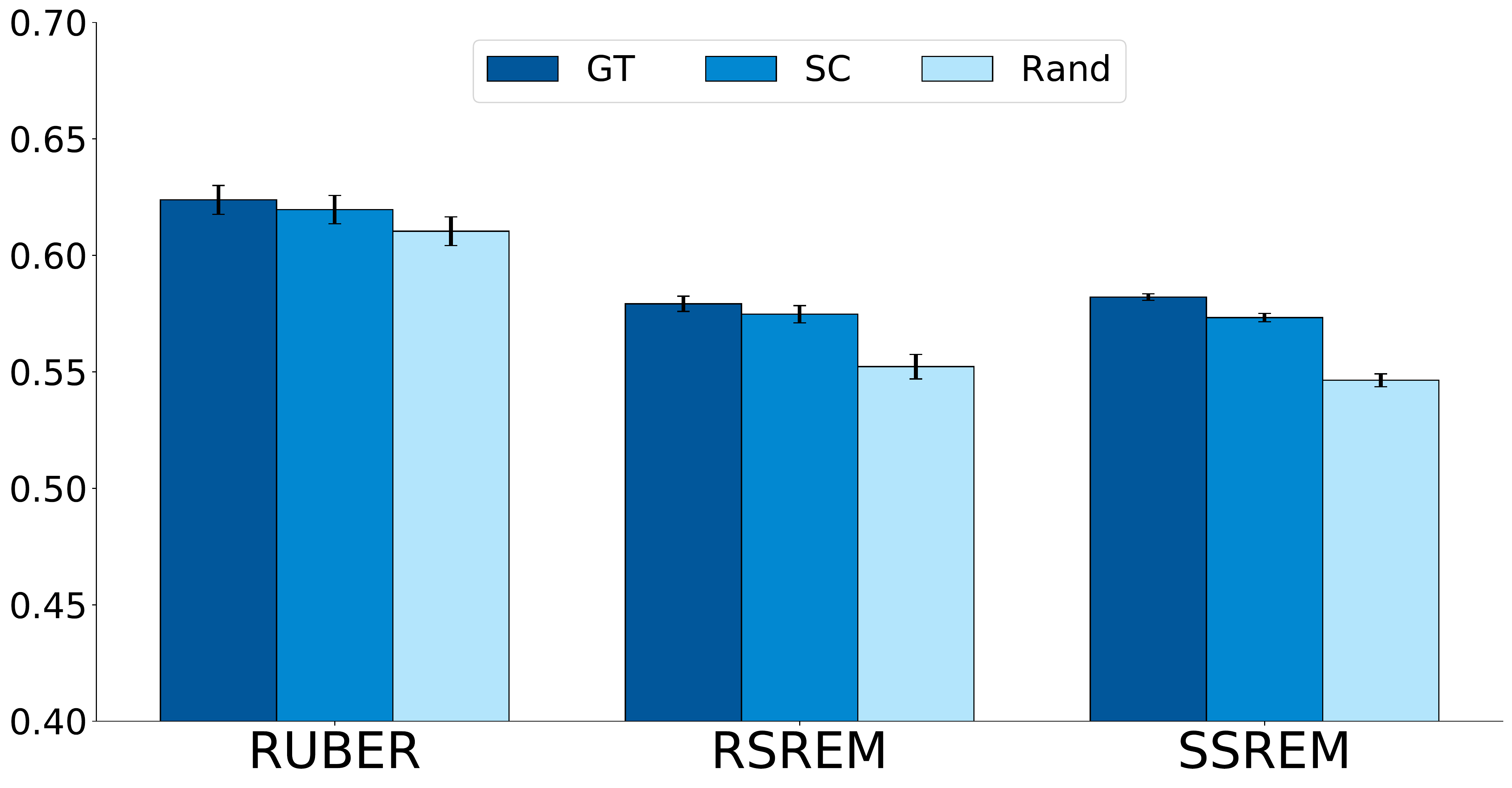}
    \caption{
        Difference of scores on various responses in Movie corpus.
        The range of the vertical error bar is a 95\% confidence interval of the values among the responses.
        SSREM outperforms the other models for identifying true and false responses.
    }
	\label{fig:metric_pre_result2}
\end{figure}

In the experiment on comparing with human scores on the movie dialogs corpus, Table \ref{tab:correlations_cornell} shows the results.
First, BLEU, ROUGE, and EMB are not correlated with human scores.
RUBER shows worse performance than testing on the Twitter corpus.
RSREM performs better than RUBER and other baselines,
but it also shows worse performance than testing on the Twitter corpus.
Finally, SSREM outperforms all other methods for two correlations with low $p$-values.
It shows the effectiveness of using speaker sensitive negative samples for the new corpus.
Figure \ref{fig:metric_human_relationship_scatter_tc} shows the similar results by plotting scatter plots.

In the experiment on identifying true and false responses with the movie dialogs corpus,
Figure \ref{fig:metric_pre_result2} shows the results of the identification task.
RUBER performs poor on distinguishing between $GT$ and $Rand$ statistically significantly.
RSREM performs better than RUBER.
And SSREM outperforms the other two models for identifying all cases in the new corpus.

\section{Conclusion and Future Work}
\label{sec:conclusion}
In this paper, we presented SSREM, an automatic evaluation model for conversational response generation.
SSREM looks at the context of the conversation and the ground-truth response together.
We proposed negative sampling with speaker sensitive samples to train SSREM.
We showed that SSREM outperforms the other metrics including RSREM that uses random negative samples only.
We also showed that SSREM is effective in evaluating a movie conversation corpus even when it is trained with Twitter conversations.

There are several future directions to improve SSREM.
First, we can make SSREM more robust on adversarial attacks. 
\newcite{sai2019reevaluating} shows limitations of ADEM on adversarial attacks such as removing stopwords and replacing words with synonyms.
We investigated another type of the adversarial attack named copy mechanism that copies one of the utterances in the context as the generated response.
All existing automatic evaluation methods including RUBER that compare the context and the response can be cheated by the copy mechanism.
SSREM is also susceptible.
However, SSREM is fooled less than other existing models because SSREM learns with negative samples from the set of utterances in the same conversation.
SSREM learns to differentiate among utterances in the same context.
We show this empirically with an experiment to identify true and false responses (Sec \ref{sec:exp2-results}).
When we look at the mean score for the context utterances that shows this copy mechanism compared to the mean score of the ground-truth response (GT), the mean score of context utterances is 0.07 higher by RUBER, but only 0.01 higher by SSREM.
SSREM does not give lower scores for the context utterances than GT, but it is not as bad as RUBER.
We will make SSREM more robust on the attacks.

Second, we can improve SSREM for a higher correlation with human judgement.
We chose to approach SSREM with a classification loss because it is simple and widely used to estimate the models using negative sampling. 
Although the classification loss is simple, SSREM outperforms all existing automatic evaluation models.
However, as Table \ref{tab:metric_neg_sets_results} and Figure \ref{fig:metric_pre_result1} are shown, each negative samples has different correlation with the context.
We will use ranking loss \cite{6909576,7298682} to learn the difference among samples.
Recently, \newcite{Zhang2020BERTScore} uses BERT \cite{devlin-etal-2019-bert} to evaluate generated candidate sentences by comparing reference sentence.
We used word embeddings to represent an utterance to the vector for the simplicity, but contextual embeddings are much better since it generates more context-related representation than word embeddings.
We will use the contextual embedding to represent utterances.

Third, we can extend using SSREM to various conversation corpora such as task-oriented dialogues.
We trained and tested SSREM on open-domain conversation corpora.
However, contextual coherence between the input context and the generated text is important in multi-turn conversations.
We will apply SSREM to various conversation tasks for evaluating the generated text automatically.
We will explore these directions in our future work.

\section*{Acknowledgments}
We would like to thank Jeongmin Byun\footnote{\url{https://jmbyun.github.io}} for building the annotation webpage, and the anonymous reviewers for helpful questions and comments.
This work was supported by Institute for Information \& communications Technology Planning \& Evaluation (IITP) grant funded by the Korea government (MSIT) (No.2017-0-01779, A machine learning and statistical inference framework for explainable artificial intelligence).

\bibliography{acl2020}
\bibliographystyle{acl_natbib}

\end{document}